\documentclass[conference]{IEEEtran}
\IEEEoverridecommandlockouts
\usepackage{cite}
\usepackage{amsmath,amssymb,amsfonts}
\usepackage{graphicx}
\usepackage{textcomp}
\usepackage{xcolor}

\usepackage{url}

\usepackage{amsmath,flushend,boldline,paralist}
\usepackage{booktabs}
\usepackage{amsfonts}
\usepackage{amsbsy}
\usepackage{bbm}
\usepackage{algorithm}
\usepackage{algorithmicx,tikz}
\usepackage{algcompatible}
\usepackage{bm}
\usepackage{enumitem}
\usepackage{lipsum}
\usepackage{multirow}
\usepackage[normalem]{ulem}

\usepackage{graphicx}
\usepackage{subcaption}

\newcommand{\name}{{\textsc{DPPIN}}}

\newtheorem{problem}{Problem}

\newcommand{\hide}[1]{}

\def\BibTeX{{\rm B\kern-.05em{\sc i\kern-.025em b}\kern-.08em
    T\kern-.1667em\lower.7ex\hbox{E}\kern-.125emX}}
\begin{document}

\title{DPPIN: A Biological Repository of Dynamic Protein-Protein Interaction Network Data}

\author{\IEEEauthorblockN{Dongqi Fu}
\IEEEauthorblockA{
\textit{University of Illinois at Urbana-Champaign}\\
Illinois, USA \\
dongqif2@illinois.edu}
\and
\IEEEauthorblockN{Jingrui He}
\IEEEauthorblockA{
\textit{University of Illinois at Urbana-Champaign}\\
Illinois, USA \\
jingrui@illinois.edu}
}

\maketitle

\begin{abstract}
In the big data era, the relationship between entries becomes more and more complex. Many graph (or network) algorithms have already paid attention to dynamic networks, which are more suitable than static ones for fitting the complex real-world scenarios with evolving structures and features.
To contribute to the dynamic network representation learning and mining research, we provide a new bunch of label-adequate, dynamics-meaningful, and attribute-sufficient dynamic networks from the health domain. To be specific, in our proposed repository \name, we totally have 12 individual dynamic network datasets at different scales, and each dataset is a dynamic protein-protein interaction network describing protein-level interactions of yeast cells. We hope these domain-specific node features, structure evolution patterns, and node and graph labels could inspire the regularization techniques to increase the performance of graph machine learning algorithms in a more complex setting.
Also, we link potential applications with our \name\ by designing various dynamic graph experiments, where \name\ could indicate future research opportunities for some tasks by presenting challenges on state-of-the-art baseline algorithms. Finally, we identify future directions to improve the utility of this repository and welcome constructive inputs from the community. All resources (e.g., data and code) of this work are deployed and publicly available at~\url{https://github.com/DongqiFu/DPPIN}.
\end{abstract}


\section{Introduction}
\label{sec:introduction}
Networks (or graphs)\footnote{We use the term "network" and "graph" interchangeably throughout the paper.} are complex data structures containing comprehensive node-attribute and node-interaction information, which attracts many research interests in network representation learning algorithms~\cite{zhang2019graph} and network mining tasks~\cite{DBLP:journals/csur/ChakrabartiF06} to serve for a wide range of real-world applications such as information retrieval~\cite{DBLP:conf/sigir/MaoXZLTH20}, recommendation~\cite{DBLP:conf/kdd/ZhouXWNKAH21}, fraud detection~\cite{DBLP:conf/sigir/LiuDYDP20}, question answering~\cite{DBLP:conf/kdd/LiuDXXT22}, time series imputation~\cite{DBLP:conf/www/JingTZ21}, and drug discovery~\cite{DBLP:conf/icml/DaiDS16}.
To fit the real-world networks evolving attributes and typologies, many graph representation learning and mining algorithms have transferred from the static setting to the dynamic setting~\cite{DBLP:journals/csur/AggarwalS14, DBLP:journals/jmlr/KazemiGJKSFP20, DBLP:conf/kdd/ZhouZ0H20, DBLP:conf/kdd/FuZH20, DBLP:conf/cikm/FuXLTH20, DBLP:conf/sigir/FuH21, DBLP:conf/kdd/FuFMTH22, DBLP:conf/cikm/ZhouZF0H22, DBLP:conf/cikm/FuBTMH22}, where the network structure (i.e., graph topology) and the node features are evolving and dependent on time.

\begin{figure}[t]
\includegraphics[width=0.49\textwidth]{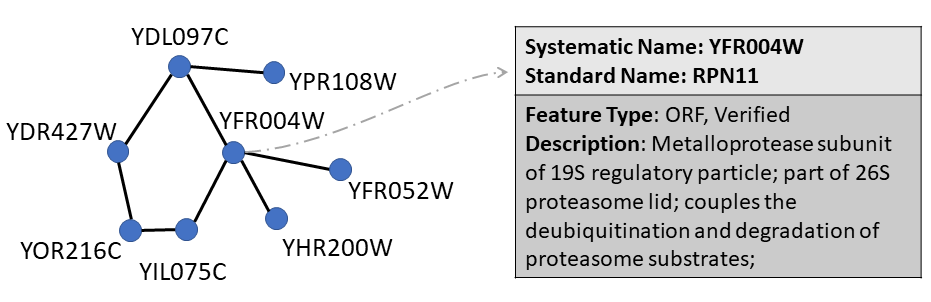}
\centering
\caption[A Subgraph Extracted from the Static Protein-Protein Interaction Network of Yeast Cells. Each node stands for a gene
coding protein, and the description of each protein node can
be extracted from the Saccharomyces Genome Database.]{A Subgraph Extracted from the Static Protein-Protein Interaction Network of Yeast Cells~\cite{babu2012interaction}. Each node stands for a gene
coding protein, and the description information of each protein node can
be retrieved from the Saccharomyces Genome Database.\footnotemark}
\label{fig:static_network}
\end{figure}

However, compared with the amount of publicly available static network datasets, the dynamic network datasets are not that sufficient. This phenomenon may hinder the related research progress. To contribute to the dynamic network representation learning and mining research community, we provide a new dynamic network repository from the biological domain, named \name. To be specific, \name\ has 12 individual dynamic network datasets at different scales describing dynamic protein-protein interactions of yeast cells, and each dynamic network dataset in \name\ is label-adequate (i.e., high label rate of nodes), dynamics-meaningful (i.e., metabolic patterns of yeast cells), and attribute-sufficient (i.e., accessible node features and edge features). We hope these domain-specific node features, structure evolution patterns, and node and graph labels could inspire the next-generation graph machine learning algorithms in the dynamic setting.

\footnotetext{\url{https://www.yeastgenome.org/}}

\begin{figure*}[t]
\includegraphics[width=0.9\textwidth]{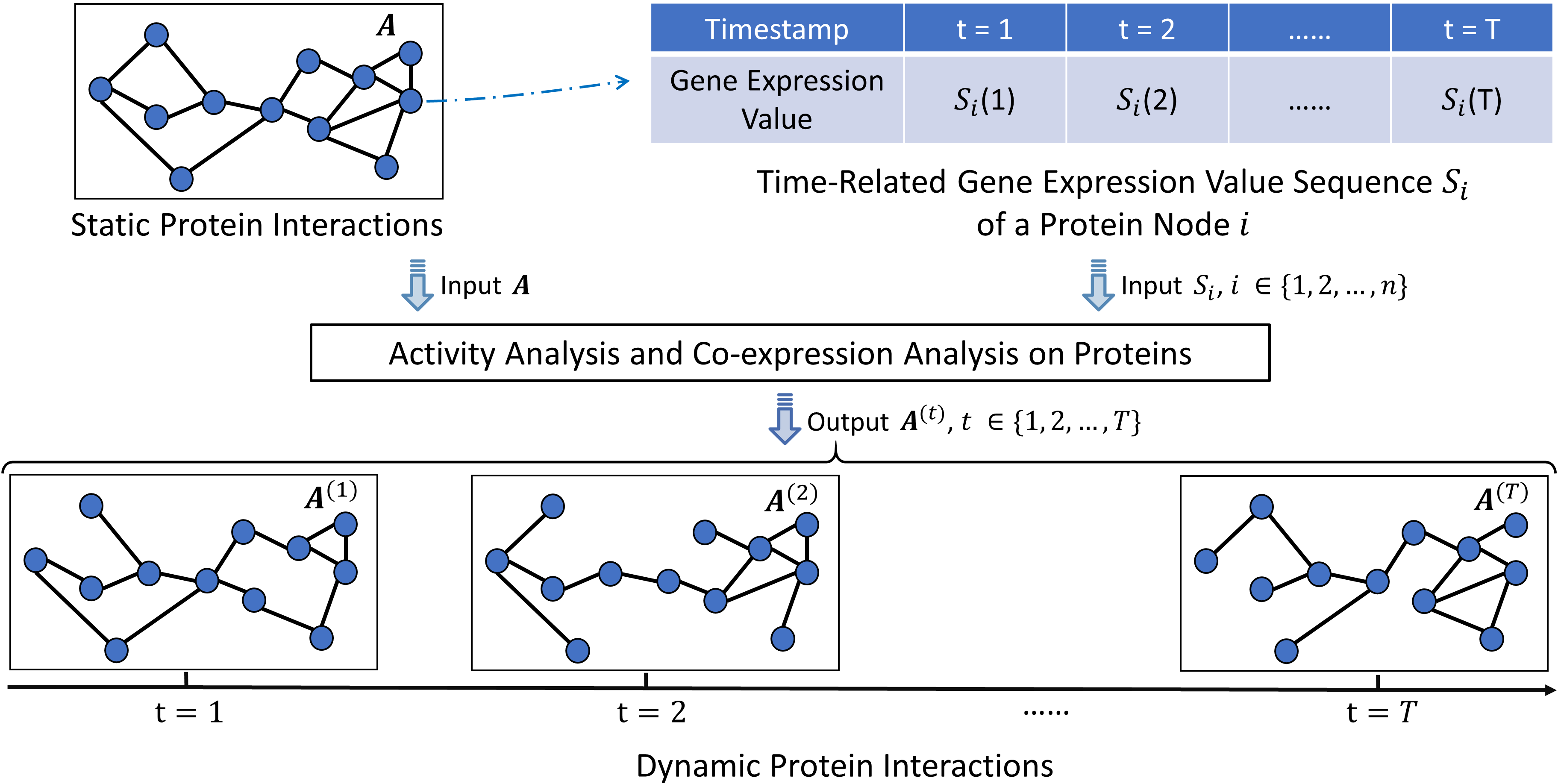}
\centering
\caption{The Dynamic Network Structure Generation Process for each Dataset in \name.}
\label{fig:generation_process}
\end{figure*}

In this paper, we first start by introducing the generation process of each dynamic network dataset in our \name, by which a static protein-protein interaction network as shown in Figure~\ref{fig:static_network} can be decomposed into different timestamps as shown in Figure~\ref{fig:generation_process}. Moreover, during the generation process, the details about the gene expression value calculation, node feature determination, and node label retrieval are discussed in Section~\ref{sec:generated_dataset} with the generation statistics. 
The utility of \name\ data repository are linked with the real-world problems by potential applications in Section~\ref{sec:potential_applications}.
With different generated network datasets in \name, we design extensive experiments such like dynamic spectral clustering and dynamic graph classification experiments in Section~\ref{sec:experiments}, where several experimental results suggest that \name\ presents challenges on state-of-the-art baseline algorithms and indicates future research opportunities.
The data source, generation process program, and other necessary information of our \name\ repository are well documented and released\footnote{\url{https://github.com/DongqiFu/DPPIN}}.

Our main contributions are summarized as follows.
\begin{itemize}[noitemsep,topsep=0pt,parsep=0pt,partopsep=0pt, leftmargin=*]
    \item \textbf{Datasets}: Regarding the biological domain, beyond many static networks\footnote{\url{https://www.inetbio.org/yeastnet/downloadnetwork.php}}, we provide a new repository of 12 dynamic protein-protein interaction network datasets to contribute to the dynamic network machine learning research community.
    \item \textbf{Experiments}: To demonstrate the utility of \name, we design extensive experiments and discuss how networks of \name\ can inspire baseline algorithms to get improved.
    \item \textbf{Future Work}: We also indicate the future work directions to improve our \name\ repository for better services.
\end{itemize}

\section{Preliminary}
\label{sec:preliminaries}
We use lowercase letters (e.g., $\alpha$) for scalars, capital letters (e.g., $V$) for sets, bold lowercase letters for column vectors (e.g., $\bm{p}$), bold capital letters for matrices (e.g., $\bm{A}$), and parenthesized superscripts to denote the timestamp of matrices (e.g., $\bm{A}^{(t)}$).

A protein-protein interaction network (shot for PPIN in the rest) is denoted as an undirected graph $G = (V, E)$, where each node represents a gene coding protein (shot for protein in the rest), each edge represents a protein-protein interaction, and matrix $\bm{A} \in \mathbb{R}^{n \times n}$ denotes the adjacency matrix encoding interactions of all nodes~\cite{rani2019detection}.
Moreover, protein-protein interactions in a cell are usually dynamic and change over time~\cite{hegde2008dynamic, chen2014identifying, shen2016mining, shen2017identifying, thanasomboon2020exploring, hu2021survey}. Many different efforts have been devoted to the dynamic PPIN construction~\cite{wang2013construction, wang2014dynamic, zhang2016method, zhang2016construction}, where a dynamic PPIN is represented as a sequence of observed snapshots, i.e., $\tilde{G} = \{G^{(t)}\}$ with $t \in \{1,2,\ldots, T\}$. Each snapshot is represented as $G^{(t)} = (V^{(t)}, E^{(t)})$ plus optional node and edge features, and $V^{(t)}$ and $E^{(t)}$ denote the nodes and edges that existed at time $t$, respectively.
Note that the number of nodes at different timestamps is different. For the clear notation, we denote $|V^{(t)}| = |V|$ in the paper, which means a node without any connections at time $t$ can just be regarded as a dangling node at that time.

To generate dynamic networks for our \name, we apply the dynamic protein-protein interaction network construction method~\cite{zhang2016method}, which could use both gene expression variance analysis and co-expression correlations analysis to establish dynamic PPINs. The generation process shown in Figure~\ref{fig:generation_process} can be described as follows. Given a static PPIN with the adjacency matrix $\bm{A} \in \mathbb{R}^{n \times n}$ and time-aware gene expression value sequences $S_{i}$ for each node $i$ at each timestamp $t \in \{1, \ldots,  T\}$, \textit{i.e., $S_{i}(t)$ denotes the gene expression value of protein node $i$ at timestamp $t$}, we sequentially build the dynamic PPIN $\tilde{G} =\{G^{(1)}, G^{(2)}, \ldots, G^{(T)}\}$, which is a sequence of snapshots $G^{(t)}$ at each observed timestamp $t$.

\section{Generated Data in \name}
\label{sec:generated_dataset}
In this section, we first introduce the dynamic structure generation process. Then, we introduce the feature and label extraction the with the corresponding biological meaning. The statistics of each dynamic network dataset in \name\ are summarized in Table~\ref{TB:dynamic_data}.

\subsection{Dynamic Structure Generation Process}
The generation process requires two inputs, as shown in Figure~\ref{fig:generation_process}, which are (1) a static PPIN and (2) time-aware gene expression values of each protein in that static PPIN.
Then the generation process is divided into three sequential steps: (1) determine active proteins at each desired timestamp, (2) determine co-expressed protein pairs at that timestamp, and (3) preserve both active and co-expressed protein connections for that timestamp snapshot. 

\textbf{Determine Active Proteins}. If the gene expression value of a protein is higher than a pre-defined threshold at a certain timestamp, then that protein is considered to be active for that timestamp. In~\cite{wang2013construction}, the active probability of proteins is expressed as follows.
\begin{equation}
    p^{(t)}(i) = 
    \begin{cases}
    0.99, &\text{if}~ S_{i}(t) \geq Th_{3}(i)\\
    0.95, &\text{if}~ Th_{3}(i) >S_{i}(t) \geq Th_{2}(i)\\
    0.68, &\text{if}~ Th_{2}(i) >S_{i}(t) \geq Th_{1}(i) \\
    0.0, &\text{if}~ S_{i}(t)< Th_{1}(i)
    \end{cases}
\label{eq:hierarchy}
\end{equation}
where $S_{i}$ is the sequential gene expression data of $i$-th protein over a period of time, and $S_{i}(t)$ denotes the expression value of the $i$-th protein at time $t$. $\bm{p}^{(t)} \in \mathbb{R}^{n \times 1}$ denotes the probability vector at time $t$, which is a column vector encoding the active probability of each protein at time $t$. Scalars $Th_{3}$, $Th_{2}$, and $Th_{1}$ are three thresholds, which are expressed as follows.
\begin{equation}
    Th_{k}(i) = \mu(S_{i}) + k \sigma(S_{i})(1-\frac{1}{1+\sigma^{2}(S_{i})}), ~k \in \{1,2,3\}
\label{eq:threshold}
\end{equation}
where $\mu(S_{i})$ stands for the mean of the entire sequence $S_{i}$, and $\sigma(S_{i})$ denotes the standard deviation of $S_{i}$.

With the computed probability vector $\bm{p}^{(t)}$ at time $t$, we can obtain the activity matrix $\bm{A}_{act}^{(t)} \in \mathbb{R}^{n \times n}$ at time $t$ as follows.
\begin{equation}
    \bm{A}_{act}^{(t)} = \bm{p}^{(t)} \bm{p}^{\top(t)}
\label{eq:activity_matrix}
\end{equation}
where $\bm{p}^{\top(t)}$ denotes the transpose of the column vector $\bm{p}^{(t)}$.

\textbf{Determine Co-expressed Protein Pairs}. The co-expression correlation coefficient is a strong indicator of protein functional associations~\cite{zhang2016method} that is used to indicate whether co-expressed genes have the same expression variance patterns across different conditions~\cite{wang2013construction}.
To investigate whether two protein nodes are co-expressed at time $t$, the co-expression matrix $\bm{A}_{coe}^{(t)} \in \mathbb{R}^{n \times n}$ of a protein network is expressed as follows with the coefficient.
\begin{equation}
    A_{coe}^{(t)}(i,j) = 
    \begin{cases}
    |PCC^{(t)}(i,j)|, &\text{if}~ |PCC^{(t)}(i,j)| \geq Th_{PCC}\\
    0, &\text{if}~ |PCC^{(t)}(i,j)| < Th_{PCC}
    \end{cases}
\label{eq:co-expression_matrix}
\end{equation}
where $PCC^{(t)}$ is the Pearson Correlation Coefficient function at time $t$ that takes $\{S_{i}(t-1), S_{i}(t), S_{i}(t+1)\}$ and $\{S_{j}(t-1), S_{j}(t), S_{j}(t+1)\}$ from two protein nodes $i$ and $j$ as the input. $Th_{PCC}$ is the pre-defined threshold for Pearson Correlation Coefficient, which is set to be 0.5 through preliminary experiments~\cite{zhang2016method}.

\textbf{Determine Active and Co-expressed Protein Interactions}. With the computed activity matrix $\bm{A}_{act}^{(t)} \in \mathbb{R}^{n \times n}$ and co-expression matrix $\bm{A}_{coe}^{(t)} \in \mathbb{R}^{n \times n}$ at time $t$, we can now indicate active and co-expressed protein interactions to form the weighted adjacency matrix at time $t$, denoted as $\bm{A}^{(t)}$.
\begin{equation}
    \bm{A}^{(t)} = \bm{A}_{act}^{(t)} \odot \bm{A}_{coe}^{(t)} \odot \bm{A}
\label{eq:dyanmic_adj}
\end{equation}
where $\odot$ stands for the element-wise multiply operation, and $\bm{A} \in \mathbb{R}^{n \times n}$ denotes the adjacency matrix of the input static PPIN.

To obtain two necessary inputs of the above-mentioned methods to generate dynamic PPIN structures, we select 12 static networks from \textit{High-Throughput Protein-Protein Interactions of yeast cells}\footnote{\url{https://www.inetbio.org/yeastnet/downloadnetwork.php}} for the matrix $\bm{A}$ in Eq.~\ref{eq:dyanmic_adj}, and the \textit{GSE3431 gene expression data}~\cite{tu2005logic}\footnote{\url{https://www.ncbi.nlm.nih.gov/geo/download/?acc=GSE3431}} for the time-aware sequence $S_{i}(t)$ in Eq.~\ref{eq:hierarchy} and Eq.~\ref{eq:threshold}. In GSE3431, each gene coding protein has 36 observed gene expression values at 36 timestamps. To be specific, those 36 timestamps consist of 3 successive metabolic cycles of yeast cells, where each cycle occupies 12 timestamped intervals, and each time interval occupies 25 minutes in the real world. Thus, we could totally have 36 timestamps for each generated dynamic network in \name.

\subsection{Feature and Label Extraction}
According to Eq.~\ref{eq:dyanmic_adj}, generated dynamic network structures in \name\ are undirected and weighted. Hence, each edge is represented as $(i, j, t, w)$, where $i$ and $j$ are two nodes, $t$ denotes the timestamp, and $w$ denotes the weight computed through Eq.~\ref{eq:dyanmic_adj} and represents the active and co-expressed probability as edge features. Moreover, the node feature of each node $i$ can be directed obtained from gene expression sequences, i.e., $S_{i}(t)$.
The label of each protein node can be accessed through the Saccharomyces Genome Database\footnote{\url{https://www.yeastgenome.org/}}. For example, the protein node YNR066C is an uncharacterized protein, YLR366W is a dubious protein, and YGR062C is a verified protein.
Now, the construction of each dataset in \name\ is completed with evolving typologies, node features, edge features, and node labels.

\begin{algorithm}[h]
\caption{Dynamic Protein Interaction Networks Construction}
\begin{algorithmic}[1]
\REQUIRE
    \STATEx static adjacency matrix $\bm{A} \in \mathbb{R}^{n \times n}$, time-aware gene expression data $S_{i}(t)$ $i \in \{1,2,\ldots n\}$, $t \in \{1,2,\ldots T\}$
\ENSURE
    \STATEx weighted adjacency matrix $\bm{A}^{(t)} \in \mathbb{R}^{n \times n}$, node feature matrix $\bm{X}^{(t)} \in \mathbb{R}^{n \times t}$, $t \in \{1,2,\ldots T\}$
    \FOR{$t = 1: T$}
        \STATEx /*Determine Active Proteins*/ 
        \STATE Determine the active probability $p^{(t)} (i)$ based on $S_{i}(t)$ with Eq.~\ref{eq:hierarchy} and Eq.~\ref{eq:threshold} for each protein node $i = 1, \ldots, n$
        \STATE Construct the activity matrix $\bm{A}_{act}^{(t)}$ based on Eq.~\ref{eq:activity_matrix}
        \STATEx /*Determine Co-expressed Protein Pairs*/ 
        \STATE Construct the co-expression matrix $\bm{A}_{coe}^{(t)}$ based on $S_{i}(t)$ with Eq.~\ref{eq:co-expression_matrix}
        \STATEx /*Preserve Active and Co-expressed Protein Pairs*/ 
        \STATE Construct the weight adjacency matrix $\bm{A}^{t}$ at time $t$ based on Eq.~\ref{eq:dyanmic_adj}
        \STATEx /*Extend Node Features via Value Concatenation*/
        \STATE $X^{(t)}(i,:) = S_{i}(1)~||~\ldots~||~S_{i}(t)$
    \ENDFOR
\end{algorithmic}
\label{Alg_1}
\end{algorithm}

\begin{table*}[t]
\caption{Selected Static PPINs for Generating Dynamic PPINs}
\centering
\begin{tabular}{|l|l|}
\hline
Static HT PPINs & Description of Protein Identification Methods \\ \hline\hline
Uetz~\cite{uetz2000comprehensive} & Genome-scale yeast-two-hybrid (Y2H) screen \\ \hline
Ito~\cite{ito2001comprehensive} & Genome-scale Y2H with pooled library \\ \hline
Ho\cite{ho2002systematic} & Genome-scale affinity purification followed by mass spectrometry analysis of co-purified proteins (APMS) \\ \hline
Gavin\cite{gavin2006proteome} & Genome-scale APMS \\ \hline
Krogan (LCMS)\cite{krogan2006global} & Genome-scale APMS with liquid chromatography tandem mass spectrometry (LCMS) \\ \hline
Krogan (MALDI)\cite{krogan2006global} & Genome-scale APMS with matrix-assisted laser desorption/ionization (MALDI) \\ \hline
Yu\cite{yu2008high} & Genome-scale Y2H \\ \hline
Breitkreutz\cite{breitkreutz2010global} & Large-scale APMS for kinase and phosphatase interactions \\ \hline
Babu\cite{babu2012interaction} & Large-scale APMS for membrane proteins \\ \hline
Lambert\cite{lambert2010defining} & Protein interactions for chromatin-related proteins \\ \hline
Tarassov\cite{tarassov2008vivo} & Genome-wide protein-protein interactions \\ \hline
Hazbun\cite{hazbun2003assigning} & Interactome for 100 essential genes \\ \hline
\end{tabular}
\label{TB:static_data}
\end{table*}

\begin{table*}[t]
\caption{Generated Dynamic Network Datasets in \name}
\centering
\scalebox{1}{
\begin{tabular}{|c|c|c|c|c|c|c|}
\hline
Generated Dynamic PPINs & $\#$Nodes & $\#$Edges  & Node Features & Edge Features & Node Label Rate&  $\#$Timestamps \\ \hline\hline
DPPIN-Uetz              & 922      & 2,159     &\checkmark &\checkmark &921/922 (99.89\%)   & 36         \\ \hline
DPPIN-Ito               & 2,856    & 8,638     &\checkmark &\checkmark &2854/2856 (99.93\%) & 36         \\ \hline
DPPIN-Ho                & 1,548    & 42,220    &\checkmark &\checkmark &1547/1548 (99.93\%) & 36         \\ \hline
DPPIN-Gavin             & 2,541    & 140,040   &\checkmark &\checkmark &2538/2541 (99.88\%) & 36         \\ \hline
DPPIN-Krogan (LCMS)     & 2,211    & 85,133    &\checkmark &\checkmark &2208/2211 (99.86\%) & 36         \\ \hline
DPPIN-Krogan (MALDI)    & 2,099    & 78,297    &\checkmark &\checkmark &2097/2099 (99.90\%) & 36         \\ \hline
DPPIN-Yu                & 1,163    & 3,602     &\checkmark &\checkmark &1160/1163 (99.74\%) & 36         \\ \hline
DPPIN-Breitkreutz       & 869      & 39,250    &\checkmark &\checkmark &869/869 (100.00\%)  & 36         \\ \hline
DPPIN-Babu              & 5,003    & 111,466   &\checkmark &\checkmark &4997/5003 (99.88\%) & 36         \\ \hline
DPPIN-Lambert           & 697      & 6,654     &\checkmark &\checkmark &697/697 (100.00\%)  & 36         \\ \hline
DPPIN-Tarassov          & 1,053    & 4,826     &\checkmark &\checkmark &1051/1053 (99.81\%) & 36         \\ \hline
DPPIN-Hazbun            & 143      & 1,959     &\checkmark &\checkmark &143/143 (100.00\%)  & 36         \\ \hline
\end{tabular}}
\label{TB:dynamic_data}
\end{table*}

The entire generation is summarized in Algorithm~\ref{Alg_1}, where Steps 2--3 are responsible for building the protein activity matrix, then Step 4 is responsible for building the protein co-expression matrix, and Step 5 is responsible for constructing the weighted adjacency matrix at that time $t$. For the feature matrix, Step 6 concatenates the protein expression values from the first time to the current time and stores this concatenation as the node feature vector. Note that the length of this node feature vector is dependent on time.

\subsection{Generation Statistics}
With the static structures, gene sequence, protein feature and label information discussed in above two subsections, we are ready to call Algorithm~\ref{Alg_1} to generate dynamic networks with time-aware node features and labels. The selected static protein networks are summarized in Table~\ref{TB:static_data}. For example, in Krogan (LCMS)~\cite{krogan2006global}, the
authors identified proteins by the liquid chromatography-tandem
mass spectrometry (i.e., LCMS). Correspondingly, the dynamic version of Krogan (LCMS) is generated through Algorithm~\ref{Alg_1}, i.e., DPPIN-Krogan (LCMS) is summarized in Table~\ref{TB:dynamic_data} with other generated dynamic network data.

\section{Potential Applications}
\label{sec:potential_applications}
In this section, we discuss graph-based applications and tasks that could leverage network data of our \name\ repository. For example, the datasets of \name\ could be used for but are not limited to dense community detection, graph querying, node similarity retrieval, node property prediction, and graph property prediction in the dynamic setting.

\subsection{Dense Community Detection}
Detecting densely connected communities (i.e., node clustering or graph partitioning) is a fundamental research problem in the graph mining research community~\cite{DBLP:conf/stoc/SpielmanT04, DBLP:conf/focs/AndersenCL06}. However, in the dynamic setting, when the graph topology updates, the previously identified clusters are outdated. To fast capture new clusters, many algorithms are proposed such like~\cite{DBLP:conf/sdm/NingXCGH07, DBLP:conf/kdd/FuZH20}.

To be specific, detecting densely connected clusters can be realized in a global view or a local view with different compactness objectives. For example, the dynamic spectral clustering~\cite{DBLP:conf/sdm/NingXCGH07} can be described as follows.

\begin{problem}{Dynamic Spectral Clustering}
\begin{description}
\item[Input:] (i) a dynamic graph $\tilde{G} = \{G^{(1)}, G^{(2)}, \ldots, G^{(T)}\}$, and (ii) the desired number of disjoint clusters $q$.
\item[Output:] $q$ disjoint clusters $\{C^{(t)}_{1}, C^{(t)}_{2}, \ldots, C^{(t)}_{q} \}$ minimizing the normalized cut, and $G^{(t)}=\bigcup_{i = 1}^{q} C^{(t)}_{i}$ at each timestamp $t \in \{1,2, \ldots ,T\}$.
\end{description}
\end{problem}

Instead of exploring the entire graph, local clustering algorithms try to identify a dense local cluster near the user-specified seed node. For example, the problem of dynamic local clustering~\cite{DBLP:conf/kdd/FuZH20} can be generalized as follows.

\begin{problem}{Dynamic Local Clustering}
\begin{description}
\item[Input:] (i) a dynamic graph $\tilde{G} = \{G^{(1)}, G^{(2)}, \ldots, G^{(T)}\}$, (ii) a seed node $u$, and (iii) the conductance upper bound $\phi$.
\item[Output:] a local cluster $C^{(t)}$ near the seed node $u$ such that the conductance score of $C^{(t)} \leq \phi$ at each timestamp $t \in \{1,2, \ldots ,T\}$.
\end{description}
\end{problem}

With the dynamic network data from our \name, dynamic clustering algorithms can be tested and further improved for faster and denser solutions.

\subsection{Graph Querying}
Searching whether a smaller specific query graph exists in a larger data graph has a very wide range of applications such as intrusion
detection, VLSI reverse engineering, and chemical compound detection~\cite{DBLP:conf/kdd/TongFGE07, DBLP:conf/edbt/ZhangLY09}. To save the computational complexity for the query process, many dynamic subgraph matching methods are proposed to leverage the historical information to fast produce the \textit{exact matching} results~\cite{DBLP:conf/sigmod/KimSHLHCSJ18, DBLP:conf/bigdataconf/LiuDXT19}, i.e., the exact positions of the query graph in the data graph. Generally speaking, the dynamic subgraph exact matching can be described as follows.

\begin{problem}{Dynamic Subgraph Exact Matching}
\begin{description}
\item[Input:] (i) a dynamic data graph $\tilde{G} = \{G^{(1)}, G^{(2)}, \ldots, G^{(T)}\}$, and (ii) a query graph $Q$.
\item[Output:] the positions (indexed by nodes) where each node and each edge of the query $Q$ is matched in the data graph $G^{(t)}$ at each timestamp $t \in \{1,2, \ldots ,T\}$.
\end{description}
\end{problem}

To obtain the exact matching solutions on dynamic data graphs, our \name\ repository plays a fundamental role by providing time-evolving node interactions.

\subsection{Node Similarity Retrieval}
Measuring the proximity (or similarity) score between nodes in a graph is very important for many graph mining research problems and paves the way for many real-world applications, such as ranking and recommendation~\cite{DBLP:conf/sdm/TongPYF08}. Due to the intrinsic dynamics of real-world networks, retrieving the similarity ranking list for every node is time-consuming. Many efforts~\cite{DBLP:conf/kdd/OhsakaMK15, DBLP:conf/kdd/ZhangLG16, DBLP:conf/www/YoonJK18} have contributed to fast and accurately tracking the node proximity in the dynamic setting, where the topology is frequently evolving over time.

Formally, the proximity tracking problem in dynamic graphs~\cite{DBLP:conf/sdm/TongPYF08} can be described as follows.

\begin{problem}{Proximity Tracking}
\begin{description}
\item[Input:] (i) a dynamic graph $\tilde{G} = \{G^{(1)}, G^{(2)}, \ldots, G^{(T)}\}$, and (ii) a set $Q$ of interest nodes.
\item[Output:] the top-$k$ most related objects of each interest node in $Q$ and their proximity (or similarity) scores at each timestamp $t \in \{1,2, \ldots ,T\}$.
\end{description}
\end{problem}

Dynamic networks in our \name\ repository enable the innovation and improvement of similarity retrieval algorithms and further link prediction applications by bringing the dynamic connections in the form of protein-protein interactions.

\subsection{Node/Graph Property Prediction}
Based on predicted properties, accurately identifying (or classifying) nodes or graphs has great importance in many domains such as drug discovery~\cite{scholkopf2004kernel, DBLP:conf/icml/DaiDS16}, molecular property prediction~\cite{DBLP:conf/nips/DuvenaudMABHAA15, DBLP:conf/icml/GilmerSRVD17}, and epidemic infectious pattern analysis~\cite{DBLP:conf/wsdm/Derr0FLAT20, DBLP:conf/sdm/OettershagenK0M20}. Recently, many efforts have tried to investigate the role of temporal dependencies in identifying the node property~\cite{DBLP:conf/iclr/XuRKKA20} and graph property~\cite{DBLP:conf/sdm/OettershagenK0M20,DBLP:conf/cikm/BeladevRKGR20} for achieving higher classification accuracy in the dynamic setting.

In general, the node-level or graph-level classification problem based on property predictions can be instanced as follows.

\begin{problem}{Dynamic Node Classification}
\begin{description}
\item[Input:] (i) a dynamic graph $\tilde{G} = \{G^{(1)}, G^{(2)}, \ldots, G^{(T)}\}$, and (ii) a node label set $Y$
\item[Output:] a representation model that could output appropriate node label predictions for the acceptable classification accuracy against the label set $Y$ at each timestamp $t \in \{1,2, \ldots ,T\}$.
\end{description}
\end{problem}

\begin{problem}{Dynamic Graph Classification}
\begin{description}
\item[Input:] (i) a dynamic graph set $\{ \tilde{G}_{1}, \tilde{G}_{2},  \ldots \tilde{G}_{N}\}$, and (ii) a graph label set $Y$
\item[Output:] a representation model that could output appropriate graph label predictions for the acceptable classification accuracy against the label set $Y$.
\end{description}
\end{problem}

Our \name\ provides 12 different classes of dynamic networks, where each network has meaningful time-dependent node connections. Therefore, \name\ can be utilized by dynamic node classification and graph classification algorithms to verify whether they could capture temporal representations to improve the performance of static baseline algorithms.

\section{Experiments}
\label{sec:experiments}
In this section, we design extensive experiments to show that network datasets from our \name\ repository could pave the way for many graph machine learning algorithms. e.g., node-level unsupervised learning and graph-level supervised learning.

\subsection{Dynamic Spectral Clustering (Motif-Aware)}
In brief, spectral clustering aims to partition the input graph into disjoint dense clusters, and different spectral clustering algorithms choose different metrics to measure the compactness of the resulting clusters.
Motif-aware spectral clustering algorithm (MSC)~\cite{benson2016higher} analyzes the eigenvalue and eigenvector of the normalized motif Laplacian matrix of the input graph and then produces clusters under the constraint of the motif conductance. A motif is a high-order connected subgraph structure, and the order of a motif stands for the number of nodes in that subgraph. For example, a third-order motif can be a triangle or a three-node line. The low motif conductance score indicates that few motifs are broken during the graph partitioning, and the resulting clusters are dense.
Although motif clustering has broad application scopes~\cite{benson2016higher}, the motif-aware spectral clustering problem on dynamic graphs remains opening. To this end, we design a novel and simple incremental solution, called MSC+T, and evaluate its performance based on different datasets of \name. To be specific, MSC+T is realized based on MSC~\cite{benson2016higher} and an eigenpairs (i.e., an eigenvalue and its corresponding eigenvector) tracking method~\cite{DBLP:conf/sdm/ChenT15}.

\begin{table}[h]
\caption{Performance of Spectral Clustering Algorithms}
\centering
\scalebox{1}{
\begin{tabular}{|c|c|c|}
\hline
\multirow{2}{*}{Methods} & \multicolumn{2}{c|}{\name-Krogan (LCMS)}           \\ \cline{2-3} 
                         & Motif Conductance & Time Consumption          \\ \hline\hline
MSC                      &  0.0000                 & 2.0276s             \\ \hline
MSC+T                    &  0.0000                 & 0.0997s              \\ \hline\hline
\multirow{2}{*}{Methods} & \multicolumn{2}{c|}{\name-Ho}                     \\ \cline{2-3} 
                         & Motif Conductance    & Time Consumption        \\ \hline\hline
MSC                      & 0.0000                  & 0.7354s              \\ \hline
MSC+T                    & 0.7500                  & 0.0469s              \\ \hline
\end{tabular}}
\label{tb:dynamic_spectral_clustering}
\end{table}

In this experiment, we select \name-Krogan (LCMS) and \name-Ho datasets and set the third-order motif (i.e., triangle) being preserved during graph cuts. The static spectral clustering algorithm (i.e., MSC) is directly executed on \name-Krogan (LCMS) and \name-Ho at timestamp $t=29$ to report the performance, and the dynamic spectral clustering method (i.e., MSC+T) receives the clustering result at $t=28$ then tracks the clustering for $t=29$. We use the motif conductance to measure the compactness performance of resulting clusters, and the lower motif conductance score dictates the fewer motif structures are broken due to the graph cuts. We partition the input graph into two clusters, and the performance of MSC and MSC+T is shown in Table~\ref{tb:dynamic_spectral_clustering}.

\textbf{Research Opportunities}. We can observe that MSC achieves the better clustering compactness but consumes a larger amount of time. Although MSC+T outputs the solution in a fast manner, the compactness of clusters is not always ideal. An intuitive explanation is that the graph structure changes (between two consecutive timestamps) beyond the representation ability of MSC+T in terms of tracking the exact eigenvalues and eigenvectors. Using the dynamic networks from \name\ identifies the future research direction that designing more dynamics-informative and accurate dynamic spectral clustering methods is necessary.

\subsection{Dynamic Local Clustering (Conductance-Guided)}
Unlike spectral clustering algorithms standing in a global optimization view, local clustering only cares about the local cluster compactness. PageRank-Nibble~\cite{DBLP:conf/focs/AndersenCL06} is a traditional local clustering algorithm designed for static graphs, which uses the stationary distribution of random walks to search an edge-preserving dense local cluster near the seed node, guaranteeing a lower bound of graph conductance. TPPR~\cite{DBLP:conf/kdd/OhsakaMK15} could track that stationary distribution of PageRank-Nibble and fast update the local cluster when a new graph structure arrives.

\textbf{Research Opportunities}. Setting the same seed node (i.e., YOL033W in \name-Krogan (MALDI) and YER052C in \name-Ho), we report the local clustering compactness performance (i.e., conductance) of PageRank-Nibble (static result at $t=28$) and TPPR (tracking result at $t=29$ from $t=28$) on two dynamic network datasets in Table~\ref{tb:dynamic_local_clustering}. Similar to the performance comparison in Table~\ref{tb:dynamic_spectral_clustering}, the dynamic local clustering method (i.e., TPPR) also fast provides the solution but does not always guarantee optimal. In this viewpoint, dynamic networks of \name\ provide the foundation for future advanced dynamic local clustering algorithms.

\begin{table}[h]
\caption{Performance of Local Clustering Algorithms}
\centering
\scalebox{1}{
\begin{tabular}{|c|c|c|}
\hline
\multirow{2}{*}{Methods} & \multicolumn{2}{c|}{\name-Krogan (MALDI)}           \\ \cline{2-3} 
                         & Conductance           & Time Consumption       \\ \hline\hline
PageRank-Nibble          & 0.3775                & 4.9119s             \\ \hline
TPPR                     & 0.7109                & 4.2645s             \\ \hline\hline
\multirow{2}{*}{Methods} & \multicolumn{2}{c|}{\name-Ho}                     \\ \cline{2-3} 
                         &  Conductance          & Time Consumption       \\ \hline\hline
PageRank-Nibble          & 0.3364                & 1.2128s             \\ \hline
TPPR                     & 0.6486                & 0.9415s             \\ \hline
\end{tabular}}
\label{tb:dynamic_local_clustering}
\end{table}

\subsection{Dynamic Subgraph Matching (Clique-Based)}
\begin{figure}[h]
    \centering
    \includegraphics[width=0.45\textwidth]{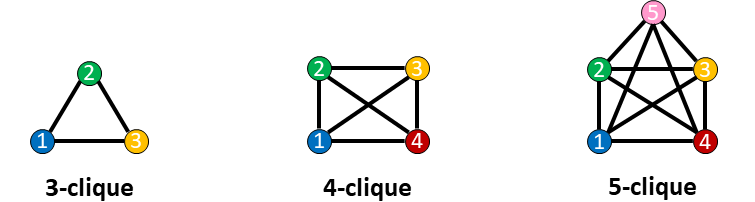}
    \caption{Examples of $k$-cliques.}
    \label{fig:cliques}
\end{figure}

Cliques are complete connected subgraphs, which means each pair of nodes is connected. A $k$-clique means that the clique structure has $k$ nodes, and every two nodes are connected. Examples of $k$-cliques (i.e., $k= \{3,4,5\}$) are shown in Figure~\ref{fig:cliques}.

\textbf{Research Opportunities}. Determining whether a data graph contains a specific clique and listing all matches has potential research value in many domains, such as identifying protein complexes and discovering new groups of functionally associated proteins~\cite{10.1093/bioinformatics/btl039}. Aiming at 3-cliques and 4-cliques, we use G-Finder~\cite{DBLP:conf/bigdataconf/LiuDXT19} to enumerate all matched subgraphs in all dynamic network datasets of \name\ at $t=28$ and $t=29$, respectively. The number of matched subgraphs is reported in Table~\ref{tb:dynamic_subgraph_matching}. We can observe that different network datasets of \name\ have very different distributions of the number of cliques. Moreover, even some network dataset has a considerably large number of nodes, but it is sparse and corresponding the number of cliques is small.

\begin{table}[h]
\caption{Number of Exact Matched Subgraphs in \name}
\centering
\scalebox{0.93}{
\begin{tabular}{|c|c|c|c|c|}
\hline
\multirow{2}{*}{Dynamic PPINs} & \multicolumn{2}{c|}{Number of 3-cliques} & \multicolumn{2}{c|}{Number of 4-cliques} \\ \cline{2-5} 
                              & $t = 28$             & $t = 29$             & $t = 28$              & $t = 29$             \\ \hline\hline
\name-Uetz                           & 0                 & 1                   &   0                  &  0                  \\ \hline
\name-Ito                            & 3                &  1                  &   0                  &   0                 \\ \hline
\name-Ho                             & 173               &  60                  & 79                    &  9                  \\ \hline
\name-Gavin                          &285               &  589                  &  184                   &  503                  \\ \hline
\name-Krogan (LCMS)                    &129              & 202                  &   46                  & 79                   \\ \hline
\name-Krogan (MALDI)                   & 557                   & 371                  & 490                    & 171                   \\ \hline
\name-Yu                             & 0                 & 0                  &  0                   & 0                  \\ \hline
\name-Breitkreutz                    &396                  & 422                  & 233                  & 244                   \\ \hline
\name-Babu                           &111                   & 269                  & 26                    & 109                   \\ \hline
\name-Lambert                        & 6                   & 1                  &  1                   & 0                   \\ \hline
\name-Tarassov                       &  3                 &   8                &  0                   & 2                   \\ \hline
\name-Hazbun                         & 7                 &  32                 &1                     &  20                  \\ \hline
\end{tabular}}
\label{tb:dynamic_subgraph_matching}
\end{table}

\subsection{Proximity Tracking (PageRank-Based)}
The goal of proximity tracking is to monitor important nodes towards the selected interest node at each observed timestamp and how the ranking of these important nodes changes as time evolves.

Here, we use the PageRank-based proximity tracking method~\cite{DBLP:conf/sdm/TongPYF08} and report the top-10 most similar proteins to the query protein YDR377W at two timestamps, $t=25$ and $t=35$, in \name-Babu dynamic network dataset. As shown in Table~\ref{tb:dynamic_proximity_tracking}, the most similar protein to YDR377W is itself, and the ranking of following similar proteins changes at different timestamps for evolving structures.

\begin{table}[h]
\caption{Top-10 Proximity Tracking Result of YDR377W}
\centering
\scalebox{1}{
\begin{tabular}{|c|c|cc|}
\hline
\multirow{2}{*}{Query Protein} & \multirow{2}{*}{Target Network}             & \multicolumn{2}{c|}{Similarity Rankings} \\ \cline{3-4} 
                              &                                             & \multicolumn{1}{c|}{$t=25$}    & $t=35$    \\ \hline\hline
\multirow{10}{*}{YDR377W}      & \multirow{10}{*}{\name-Babu} & \multicolumn{1}{c|}{YDR377W}   & YDR377W   \\ \cline{3-4} 
                              &                                             & \multicolumn{1}{c|}{YBL099W}   & YMR001C   \\ \cline{3-4} 
                              &                                             & \multicolumn{1}{c|}{YPL271W}   & YDR291W   \\ \cline{3-4} 
                              &                                             & \multicolumn{1}{c|}{YNL225C}   & YJR090C   \\ \cline{3-4} 
                              &                                             & \multicolumn{1}{c|}{YDR126W}   & YER093C   \\ \cline{3-4} 
                              &                                             & \multicolumn{1}{c|}{YIL026C}   & YOR032C   \\ \cline{3-4} 
                              &                                             & \multicolumn{1}{c|}{YNR006W}   & YOR307C   \\ \cline{3-4} 
                              &                                             & \multicolumn{1}{c|}{YHR155W}   & YDR408C   \\ \cline{3-4} 
                              &                                             & \multicolumn{1}{c|}{YGL263W}   & YNL147W   \\ \cline{3-4} 
                              &                                             & \multicolumn{1}{c|}{YGR206W}   & YHR166C   \\ \hline
\end{tabular}}
\label{tb:dynamic_proximity_tracking}
\end{table}

\subsection{Dynamic Node Classification (Attention-Aware)}
To identify the node property (e.g., class labels), inspired by Graph Attention Network (GAT)~\cite{DBLP:conf/iclr/VelickovicCCRLB18}, Temporal Graph Attention Network (TGAT)~\cite{DBLP:conf/iclr/XuRKKA20} is a recently proposed graph neural network model for aggregating the temporal-topological neighborhood information into the node representation vector (through attention mechanisms) for improving the node classification accuracy in the dynamic setting.

\begin{table}[h]
\caption{Performance of Graph Neural Networks w.r.t. Nodel-level Classification Accuracy}
\centering
\scalebox{1}{
\begin{tabular}{|c|c|c|}
\hline
\multirow{2}{*}{Methods} & \multicolumn{2}{c|}{\name-Ito}                   \\ \cline{2-3} 
                         & Training Acc.           & Testing Acc.           \\ \hline\hline
GAT                      & 0.9426 $\pm$ 0.0004     & 0.9435 $\pm$ 0.0078    \\ \hline
TGAT                     & 0.9478 $\pm$ 0.0005     & 0.9558 $\pm$ 0.0096    \\ \hline
\end{tabular}}
\label{tb:dynamic_node_classification}
\end{table}

Here, we want to investigate whether TGAT could achieve higher node classification than GAT in our \name-Ito dataset by capturing the temporal metabolic evolution information. Based on the chronological order, we split the first $70\%$ temporal edges as the training set, the middle $15\%$ temporal edges as the validation set, and the last $15\%$ temporal edges as the testing set. Then, we report the average node classification accuracy and the standard deviation of GAT and TGAT with the varying number of test samples in Table~\ref{tb:dynamic_node_classification}, where GAT and TGAT are converged under the same condition (e.g., the same number of layers, same dropout probability, same learning rate, and same training epochs). Moreover, to make the static method GAT could take the dynamic network \name-Ito as input, we use Reduced Graph Representation~\cite{DBLP:conf/sdm/OettershagenK0M20} to map temporal graphs into dynamics-preserving static graphs. In Table~\ref{tb:dynamic_node_classification}, we can observe that TGAT achieves the higher training accuracy and testing accuracy in terms of the node classification, which suggests that temporal metabolic patterns are helpful in regularizing the representation learning process to help determine the category of nodes, TGAT could capture those temporal metabolic patterns in \name-Ito to help identify protein properties.

\subsection{Dynamic Graph Classification (Few-Shot Learning)}
To a large extent, accurately classifying different class dynamic graphs and saving labeling workload requires corresponding graph embedding algorithms to encode the class-distinctive graph property (i.e., structure and/or feature evolution pattern) appropriately in the learning process.

Here, we conduct the dynamic graph classification experiment on state-of-the-art baseline algorithms and see if datasets in \name\ could help indicate some latent research opportunities.
According to~\cite{DBLP:conf/cikm/BeladevRKGR20}, tdGraphEmbed is the first graph-level dynamic graph representation learning algorithm, which could capture the representation of each observed snapshot. We use tdGraphEmbed and call the sum pooling function to aggregate snapshot representations for the entire dynamic graph representation vector for further classification. We also involve two static graph-level representation learning algorithms, Graph2Vec~\cite{DBLP:journals/corr/NarayananCVCLJ17} and GL2Vec~\cite{DBLP:conf/iconip/ChenK19}. Furthermore, to make static algorithms deal with dynamic inputs, we use Reduced Graph Representation~\cite{DBLP:conf/sdm/OettershagenK0M20} to map evolving graphs into dynamics-preserving static graphs.
In our \name, there are 12 graph classes (indicated by network names) in total, and each class has one large dynamic graph. In each class, we take subgraphs as samples. To be specific, for each graph, we take a subgraph by extracting a length of 5 timestamps with every 3 timestamps elapsed. For example, $t_{1}$, $t_{2}$, $t_{3}$, $t_{4}$, $t_{5}$ together compose the first dynamic subgraph, and $t_{4}$, $t_{5}$, $t_{6}$, $t_{7}$, $t_{8}$ together compose the second dynamic subgraph. The extracted subgraph shares the class label with its original entire graph, and we have 11 dynamic subgraphs per class.
Therefore, we can sample subgraphs and form the $N$-way $k$-shot classification setting (i.e., $N$ classes and $k$ samples per class during meta-training and no shared class labels between the meta-training and meta-testing stages) with two few-shot classifiers: ProtoNet and its special cases kNN~\cite{DBLP:conf/nips/SnellSZ17}. We split 8 classes for the meta-training and 4 classes for the meta-testing, and we randomly shuffle this split 4 times to report the testing accuracy of the meta-testing stage in Table~\ref{tb:dynamic_graph_classification}.

\begin{table}[h]
\caption{Graph-level Classification Accuracy during Meta-testing with Varying Few-shot Settings}
\centering
\scalebox{1}{
\begin{tabular}{|c|c|c|}
\hline
\multirow{2}{*}{Methods} & \multicolumn{2}{c|}{Few-Shot Setting}     \\ \cline{2-3} 
                         & 3 way - 5 shot      & 3 way - 3 shot      \\ \hline\hline
Graph2Vec + kNN\footnotemark             & -- & -- \\ \hline
GL2Vec + kNN              &$0.0717\pm0.0900$  &$0.0917\pm0.0793$ \\ \hline
tdGraphEmbed + kNN       &$0.2167\pm0.1736$  &$0.1056\pm0.0814$ \\ \hline
Graph2Vec + ProtoNet        &$0.3792\pm0.0459$  &$0.3958\pm0.0731$ \\ \hline
GL2Vec + ProtoNet        &0.7100 $\pm$ 0.0361  &0.6625 $\pm$ 0.0407 \\ \hline
tdGraphEmbed + ProtoNet  & 0.6562 $\pm$ 0.1882 & 0.6791 $\pm$ 0.1141 \\ \hline
\end{tabular}}
\label{tb:dynamic_graph_classification}
\end{table}
\footnotetext{Cannot get the results within 48 hours on a Linux machine with a single NVIDIA Tesla V100 32GB GPU} 

\textbf{Research Opportunities}.
As shown in Table~\ref{tb:dynamic_graph_classification}, tdGraphEmbed+ProtoNet achieves the best performance in both few-shot settings. An intuitive explanation is that, compared just simply calling Reduced Graph Representation~\cite{DBLP:conf/sdm/OettershagenK0M20} for static algorithms, the dynamics representation learning process of tdGraphEmbed is more suitable for capturing the bioinformatic evolution patterns. However, tdGraphEmbed takes each snapshot individually, and we just call sum pooling to aggregate all of them. But there is still an opening question about how to aggregate each snapshot representation more reasonably. For instance, if there are some snapshots globally shared by different class dynamic graphs (i.e., different class graphs share several same or similar snapshots), then the simply summarized graph-level representation may not be class-distinctive. Based on this observation, we know that capturing the lifelong evolution pattern may be an essential factor in designing the next-generation dynamic graph-level representation learning algorithms. Some hints may be found in applying the attention mechanism~\cite{DBLP:conf/nips/VaswaniSPUJGKP17} of current node-level representation learning methods~\cite{DBLP:conf/iclr/XuRKKA20}.

\section{Future Work}
\label{sec:future_work}
Although every gene coding protein that appeared in our \name\ repository has observed time-aware expression values based on the GSE3431 gene expression data~\cite{tu2005logic}, some protein nodes are currently lacking meaningful labels, as shown in the Saccharomyces Genome Database\footnote{\url{https://www.yeastgenome.org/}}. Consequently, some dynamic network datasets in our \name\ repository have no $100.00\%$ label rate, which can be seen in Table~\ref{TB:dynamic_data}.
This defect may hinder executing supervised learning algorithms on the datasets of \name, such as some node classification tasks. In the future, we will continually follow related research of unlabeled proteins of our \name\ repository and incrementally update the label set of \name. Also, facing imperfect labels, another possibility is using self-supervised learning techniques on graphs~\cite{DBLP:conf/cikm/JingFXCCT22, DBLP:journals/corr/abs-2206-00006}.

\section*{Acknowledgement}
This work is supported by National Science Foundation under Award No. IIS-1947203, IIS-2117902, and IIS-2137468. The views and conclusions are those of the authors and should not be interpreted as representing the official policies of the funding agencies or the government.

\section{Conclusion}
\label{sec:conclusion}
In this paper, we first release a new repository of dynamic network datasets from the bioinformatics domain, called \name. To be specific, each network of \name\ consists of dynamic protein-protein interactions generated by analyzing active and co-expressed protein pairs at corresponding timestamps. Then, we connect our \name\ repository with the real-world applications that could take advantage of it through corresponding experiments, such as dense community detection, graph querying, node similarity retrieval, and node/graph property prediction in the dynamic setting. Finally, we indicate the future direction to continually improve the utility of our \name\ data repository. We hope the various evolution patterns in each network dataset of the \name\ repository could provide guidance or constraint for developing more effective dynamic graph machine learning and mining algorithms.

\bibliographystyle{plain}
\bibliography{reference.bib}

\end{document}